\documentclass[a4paper,11pt]{article}

\usepackage{amsmath,amssymb,amsfonts}
\usepackage{algorithmic}

\usepackage{textcomp}
\usepackage{xcolor}
\usepackage{todonotes}
\usepackage{booktabs}

\usepackage{fourier}
\usepackage{color}
 \usepackage{graphicx}
\usepackage{url}
\usepackage[affil-it]{authblk}
\usepackage{wrapfig}

\usepackage[T1]{fontenc}
\usepackage{times}

\begin{document}

\title{Domain Generalisation with Bidirectional Encoder Representations from Vision Transformers}

\author{Hamza Riaz and Alan F. Smeaton}
\affil{Dublin City University, Glasnevin, Dublin 9, Ireland\\
alan.smeaton@dcu.ie}
\date{}
\maketitle
\thispagestyle{empty}

\begin{abstract}
Domain generalisation involves pooling knowledge from  source domain(s) into a single model that can generalise  to unseen target domain(s). 
Recent research in domain generalisation has faced  challenges when using deep learning models as they interact with  data distributions which differ from those  they are trained on. 
Here we perform domain generalisation  on out-of-distribution (OOD) vision benchmarks using vision transformers. 
Initially we examine  four   vision transformer architectures namely ViT, LeViT, DeiT, and BEIT on out-of-distribution data.
As the bidirectional encoder representation from image transformers (BEIT) architecture performs best, we use it in further experiments on three benchmarks PACS, Home-Office and DomainNet. Our results  show  significant improvements in validation and test accuracy  and our implementation significantly overcomes  gaps between within-distribution and OOD data. 

\end{abstract}

%%%%%%%%%%%%%%%%%%%%%%
\section{Introduction}
Machine learning algorithms rely on training data distributions though in certain scenarios computer vision can  fail to generalise 
when applied to out-of-distribution (OOD) data.  
Real-world systems like autonomous vehicles can show a  decline in performance when interacting with even partially different conditions and settings  compared to their training data distributions. 

Domain generalisation is influenced by three factors: dataset types, network architectures, and model selection criteria. To overcome OOD  challenges, much work has been done including  solutions like additional data collection for different domains, adversarial learning, and data augmentation for  learning generalised invariances from the training domain 
\cite{akuzawa2020adversarial}. 
A range of pointers from the literature encourages us to explore vision transformers for domain generalisation in computer vision. We implemented a pipeline to determine the OOD capability of four available pre-trained vision transformers. Originally each  were pre-trained and  fine-tuned on ImageNet-21k and ImageNet1k respectively. Using these, we  run  inference on unseen benchmarks   including  ImageNet-Sketch, ImageNet-R (endition), Imagenet Adversarial, and Imagenet Corrupted. 
From these results, we choose BEIT for further analysis and we  fine-tune three separate models on three popular benchmarks for domain generalisation namely PACS, Home-Office and DomainNet.

\section{Vision Transformers}

Vision transformers are inherently more appropriate for domain generalisation compared to other CNNs because of factors like global understanding, handling variable-length inputs, fewer parameters, an attention mechanism, and pre-training.
A vision transformer uses the transformer architecture to analyse images for various downstream  tasks. A simple transformer architecture was initially proposed for natural language processing tasks in \cite{wolf2020transformers} which was  extended into  the vision transformer in order to handle image data \cite{dosovitskiy2020image}. The main innovation in the vision transformer is its ability to process an entire image as a sequence of patches rather than as a grid of pixels. 
Vision transformers use self-attention during the learning process and an attention score is computed by the product of query-key terms in the last layer. 

\section{OOD Inference Experiments with Pre-trained Vision Transformers}

To conduct initial OOD experiments  we  use pre-trained weights from 4  baseline vision transformers namely ViT, LeViT, DeiT and BEIT. These  are fine-tuned on ImageNet 2012 1K classes with 224×224 input resolution and 16×16 patch size except LeViT which has 256×256 input resolution. For  performance of these  on OOD datasets, we use variations of ImageNet as OOD examples namely ImageNet Sketch, ImageNet-R(edition), Imagenet Adversarial, and Imagenet Corrupted. The ImageNet-Sketch dataset has 50,000 images with 1K classes, 50 images for each of the 1,000 ImageNet classes. ImageNet-R contains 30,000 image renditions for 200 ImageNet classes which is a subset of ImageNet-1K. ImageNet-adversarial consists of adversarially filtered real-world images to fool ImageNet classifiers and it also contains 200 classes as a subset of ImageNet-1K. Finally Imagenet Corrupted consist of images with 75 common visual distractions and the goal was to improve and evaluate the robustness of models, it has 1,000 classes.

\begin{table*}[!ht]
\caption{Results using 4 vision transformers (rows) on 4 OOD related benchmarks (columns)} % title of Table
\centering % used for centering table
%\small
\resizebox{0.8\columnwidth}{!}{%
\begin{tabular}{lcccccccc}\toprule& \multicolumn{2}{c}{ImageNet-Sketch} & \multicolumn{2}{c}{ImageNet-R(edition)}& \multicolumn{2}{c}{Imagenet Adversarial} & \multicolumn{2}{c}{Imagenet Corrupted}
\\%\cmidrule(lr){2-3}\cmidrule(lr){4-5}
\midrule
Models           & Top1 Acc   &  Top5 Acc & Top1 Acc   &  Top5 Acc& Top1 Acc   &  Top5 Acc& Top1 Acc   &  Top5 Acc \\
\midrule
ViT    &35.43& 57.29 &32.82& 47.54 &12.97& 30.04& 78.06& 94.43   \\
LeViT  & 0.95 & 0.72 &0.81 & 0.44 & 9.13 & 27.14 & 73.67 & 90.98   \\
DeiT & 32.58 & 50.21 & 31.04 & 44.42 & 9.97 & 24.31 & 77.95 & 92.56  \\
\color{blue}BEIT   &\color{blue}47.55 & \color{blue}71.01 & \color{blue}44.72 & \color{blue}62.13 &\color{blue} 22.60 &\color{blue} 47.74 & \color{blue}81.88 &\color{blue} 96.41  \\% inserts single horizontal line
\bottomrule
\label{tab2} % is used to refer this table in the text
\end{tabular}
}
\end{table*}

Table~\ref{tab2}  presents the top-1 and top-5 accuracy figures for 4 selected transformers on 4 OOD-related benchmarks. Results indicate that BEIT outperforms given transformers with a notable improvement in evaluation metrics for each benchmark. The main reason BEIT surpasses others are its properties including Mask Image Modeling (MIM) with self-supervised learning of large models, a self-attention mechanism, and denoising of corrupted inputs. Thus we selected BEIT for analysis of domain generalisation benchmarks. The next section presents the methodology of our approach.

\section{Domain Generalisation Experiments}

The BEIT vision transformer was applied to the PACS, Office-Home, and DomainNet benchmarks to  test the domain generalisation capability of BEIT for small, medium and large datasets. PACS has 9,991 images with 4 domains and 7 classes which is a comparatively  smaller dataset. Office-Home has 15,588 images also with 4 domains but it has 65 classes.  DomainNet is one of the largest benchmarks for domain generalisation with more than 0.5 million images, 6 domains and 365 classes. Although fine-tuning of any vision transformer is a relatively less time-consuming process than pre-training from scratch, this also depends on the size of the dataset. For instance, PACS and Office-Home take almost 4-6 hours for fine-tuning but in the case of DomainNet our model takes almost 3 days. 
During the training  and validation steps, pre-processing includes image resizing, random horizontal flip, and normalisation. Similarly, testing includes re-sizing, centre cropping, and normalisation. 

Inspired by the work in \cite{bao2022beit}, we used the based version of BEIT transformer which has 12 transformer layers with 768 hidden  and 3,072 feed-forward networks. Each attention layer has 12 attention heads of size 64 and these are responsible for learning self-attention masks.  Each image was divided into 14*14 patches of 16*16 pixels. BEIT is trained with 8,192 visual tokens.
The version of pre-trained weights which we used were pre-trained and fine-tuned on ImageNet 21k. 

\section{Experimental Results on OOD Benchmarks}

Following fine-tuning of hyperparameters for OOD datasets, we applied well-trained and shallow networks of BEIT on the unseen testing sets from each benchmark using \url{https://github.com/huggingface/transformers}.

\begin{table*}[htbp]
\caption{Results of BEIT fine-tuning experiments on three benchmarks -- PACS, Office-Home, DomainNet}
\centering
\resizebox{0.9\columnwidth}{!}{%
\begin{tabular}{lcccccc}
\toprule
%\textbf{Table}&\multicolumn{3}{|c|}{\textbf{Table Column Head}} \\
\textbf{Benchmarks} & \textbf{\textit{Validation Top1 Acc}}& \textbf{\textit{Target Top1 Acc}}& \textbf{\textit{Validation Top5 Acc}}& \textbf{\textit{target Top5 Acc}}& \textbf{\textit{Gap}}& \textbf{\textit{Precision}} \\
\midrule
PACS & 0.96 & 0.94 & 1.0 & 0.9980 & 0.02 & 0.9464 \\
Office-Home &  0.8597 & 0.8691 & 0.9948 & 0.9679 &-0.0094 & 0.8754\\
DomainNet & 0.7019 & 0.6978 & 0.9347 & 0.8793 & 0.0041 & 0.7111\\
\bottomrule
\multicolumn{5}{l}{}
\end{tabular}
\label{tab3}
}
\end{table*}

Table~\ref{tab3} presents  results of the vision transformer-based domain generalised model including the  top-1 and top-5 scores for validation and target/test data distributions. Ideally, in domain generalisation, the gap is the difference in performance metrics like accuracy, loss or precision for Independent and Identically Distributed (IID) and for Out Of Domain (OOD) test data.  Here we consider the validation distribution as  IID and the target distribution as OOD and the gap is the difference between these.  For all three benchmarks, our vision transformer-based model  shows state-of-the-art performance. In the case of PACS, it has 0.94  accuracy and the gap is only 0.02, a sign of good domain generalisation. For Office-Home, the OOD or target accuracy is higher than the validation accuracy of IID which made the gap negative. Although our model has relatively low performance for DomainNet and Office-Home compared to PACS  the gap remains small which means the generalisation  performs effectively.

\begin{table*}[htbp]
\caption{BEIT accuracy and loss for PACS, Office-Home, and DomainNet for each domain independently}
\centering
\resizebox{0.75\columnwidth}{!}{%
\begin{tabular}{lccccccc}
\hline 
%\textbf{Table}&\multicolumn{3}{|c|}{\textbf{Table Column Head}} \\
\textbf{PACS} & \textbf{\textit{Photos}}& \textbf{\textit{Artwork}}& \textbf{\textit{Cartoon}}& \textbf{\textit{Sketch}}&  \\
\hline

Accuracy & 0.9766 & 0.9183 & 0.9578 & 0.9371 & \\
Loss &  0.0493 & 0.2507 & 0.1206 & 0.2227 & \\
\hline
\textbf{Office-Home} & \textbf{Art} & \textbf{Clipart} &\textbf{Product}  &\textbf{Real World}  & \\
\hline
Accuracy & 0.7979 & 0.8488 & 0.9324 & 0.8645 & \\
Loss & 0.7443 & 0.5947 & 0.2502 & 0.5339 & \\

\hline
\textbf{DomainNet} & \textbf{Clipart} & \textbf{Infograph} & \textbf{Painting} &\textbf{Quickdraw} & \textbf{Real World} & \textbf{Sketch} & \\
\hline
Accuracy & 0.7822 & 0.3812 & 0.6893 & 0.6727 & 0.8073 & 0.6764 & \\
Loss & 0.9129 & 3.0639 & 1.4036 & 1.2023 & 0.7915 & 1.4661 & \\
\hline
\multicolumn{5}{l}{}
\end{tabular}
\label{tab4}
}
\end{table*}

Table~\ref{tab4} presents loss and accuracy metrics for each domain. 
The first rows  present the performance of BEIT-PACS which has 4 domains namely photos, artwork, cartoon, and sketch. It is clear that the model has good performance  for the samples of photos but lower accuracy and loss for samples related to the artwork domain. The rows in Table~\ref{tab4} show performance of  BEIT-Office-Home which also has 4  domains namely art, clipart, product, and real world. Like BEIT-PACS, BEIT-Office-Home also shows lower scores for the artwork domain,  possibly because pre-trained weights do not have the high-level features because not enough artwork images were used in training. %which could be important for the artwork domain. 
In the case of BEIT-DomainNet, the model has 6 domains, 365 classes and more than 0.5 million samples from across different domains. The model has poorer performance for samples of infographs as it is a  different domain to the others. 

\begin{table*}[ht]
\caption{Comparison between our trained model and other state-of-the-art methods for OOD generalisation } % title of Table
\centering % used for centering table
\resizebox{\columnwidth}{!}{%
%\small
\begin{tabular}{lccccccc}\toprule& \multicolumn{3}{c}{PACS} & \multicolumn{3}{c}{Office-Home}& \multicolumn{1}{c}{DomainNet}
\\%\cmidrule(lr){2-3}\cmidrule(lr){4-5}
\hline
Models           & IID Accuracy   &  OOD Accuracy & Gap   &  IID Accuracy & OOD Accuracy   &  Gap & Target Accuracy \\\hline
GroupDRO& 0.95 & 0.73 & 0.22 & 0.82 & 0.52 & 0.30 & 0.337   \\
ANDMask & 0.95 & 0.72 &0.23 & 0.81 & 0.44 & 0.37 & *  \\
Mixup   & 0.97 & 0.72 & 0.25 & 0.83 & 0.53&0.30&  0.396\\
MMD     & 0.94 & 0.69 & 0.25 & 0.82 & 0.52 &0.30 &  0.394\\
DANN    & 0.94 & 0.73 &0.21 & 0.83 & 0.51 &0.32&  0.384\\
CORAL   & 0.95 & 0.77 &0.18 & 0.84 & 0.55&0.29&  0.418\\
VREx    & 0.97 & 0.80 &0.17 & 0.76 & 0.49&0.27&  0.336\\
RSC     & 0.97 & 0.77 & 0.20 & 0.83 & 0.50 &0.33 & 0.389 \\
ERM     & 0.97 & 0.78 &0.19 & 0.84 & 0.57 &0.27&  0.412\\
 
\color{blue}Our approch  &\color{blue}0.96 & \color{blue}0.94 & \color{blue}0.02 & \color{blue}0.86 &\color{blue} 0.87 &\color{blue} -0.0094 & \color{blue}0.70 \\% inserts single horizontal line
\label{tab5} % is used to refer this table in the text
\end{tabular}
}
\end{table*}

To compare our method with the state-of-the-art, Table~\ref{tab5} presents various CNN-based domain generalised algorithms using performance figures taken from \cite{riaz2023vision},\cite{gulrajani2021in}. 
The columns  present IID accuracy, OOD accuracy, and the ensuing gap, for PACS and for Office-Home. For DomainNet we consider target accuracy only. It is clear that our method substantially outperforms  state-of-the-art approaches in all benchmarks. 
If we examine the figures for PACS, our method obtains overall 0.96 and 0.94 IID and OOD accuracy respectively and the gap shrinks  to 0.02. Our method also shows similar performance for Office-Home with a negative because our model perform slightly better on OOD than IID which is a good sign for domain generalisation. Table~\ref{tab5} also shows that overall  target accuracy is not  high for all approaches but  our method still outperforms all existing approaches.
%with a difference of 37.98\%\ if we consider 31.8\%\ as average accuracy.  

\section{Conclusions}

We present an investigation into vision transformers for domain generalisation in computer vision. We fine-tuned a base version of the BEIT model for domain generalisation benchmarks  and  investigated the generalisation of other pre-trained vision transformers on OOD versions of ImageNet. 
A per-domain analysis and  detailed comparison with other  domain generalisation algorithms showed  our model yields state-of-the-art results or better for all benchmarks  significantly reducing the gaps between IID and OOD scores  in all  benchmarks. 
\\

\noindent 
{\bf Acknowledgements:} HR is funded by Science Foundation Ireland under the ML-Labs SFI Centre for Researcher Training in Machine Learning (18/CRT/6183) and AS is part-funded under  SFI/12/RC/2289\_P2.

\bibliographystyle{apalike}

\bibliography{bibliography}

\end{document}